\pdfoutput=1

\documentclass[11pt]{article}

\usepackage{ACL2023}

\usepackage{times}
\usepackage{latexsym}

\usepackage[T1]{fontenc}

\usepackage[utf8]{inputenc}

\usepackage{microtype}
\usepackage{booktabs}

\usepackage{inconsolata}
\usepackage{cleveref}
\crefformat{section}{\S#2#1#3}
\crefformat{subsection}{\S#2#1#3}
\crefformat{subsubsection}{\S#2#1#3}
\usepackage{multirow}
\usepackage{graphicx}
\title{xPQA: Cross-Lingual Product Question Answering across 12 Languages}

\author{Xiaoyu Shen$^{1}$, Akari Asai$^{2}$, \textbf{Bill Byrne}$^{1,3}$ and \textbf{Adrià de Gispert}$^{1}$ \\
$^1$Amazon Alexa AI, $^2$Univeristy of Washington, $^3$University of Cambridge\\
\tt{\{gyouu, willbyrn, agispert\}@amazon.com}}

\begin{document}
\maketitle

\begin{abstract}
Product Question Answering (PQA) systems are key in e-commerce applications to provide responses to customers’ questions as they shop for products. While existing work on PQA focuses mainly on English, in practice there is need to support multiple customer languages while leveraging product information available in English. To study this practical industrial task, we present xPQA, a large-scale annotated cross-lingual PQA dataset in 12 languages across 9 branches, and report results in (1) candidate ranking, to select the best English candidate containing the information to answer a non-English question; and (2) answer generation, to generate a natural-sounding non-English answer based on the selected English candidate.
We evaluate various approaches involving machine translation at runtime or offline, leveraging multilingual pre-trained LMs, and including or excluding xPQA training data. We find that (1) In-domain data is essential as cross-lingual rankers trained on other domains perform poorly on the PQA task; (2) Candidate ranking often prefers runtime-translation approaches while answer generation prefers multilingual approaches; (3) Translating offline to augment multilingual models helps candidate ranking mainly on languages with non-Latin scripts; and helps answer generation mainly on languages with Latin scripts.
Still, there remains a significant performance gap between the English and the cross-lingual test sets.\footnote{The xPQA dataset is released under \url{https://github.com/amazon-science/contextual-product-qa/} for research purposes.}
\end{abstract}

\section{Introduction}
Product question answering (PQA) is a key technology in e-commerce applications. 
Given a question about a product, a PQA system searches the product webpage and provides an instant answer, so that customers do not need to traverse the page by themselves or seek help from  humans~\cite{li2017alime,carmel2018product}. In our globalized world, it is essential to enable this technology for customers from different backgrounds. However, existing research focuses predominantly on English and leaves aside other language users. 
One of the biggest obstacles is the lack of datasets, which prevents us from training, evaluating and developing non-English PQA systems. Despite the growing number of multilingual QA datasets, their main focus is on general domains such as Wikipedia, which generalize poorly when applied to the PQA task, as we show in our experiments.

\begin{figure}[t]
    \centering
    \includegraphics[width=\columnwidth]{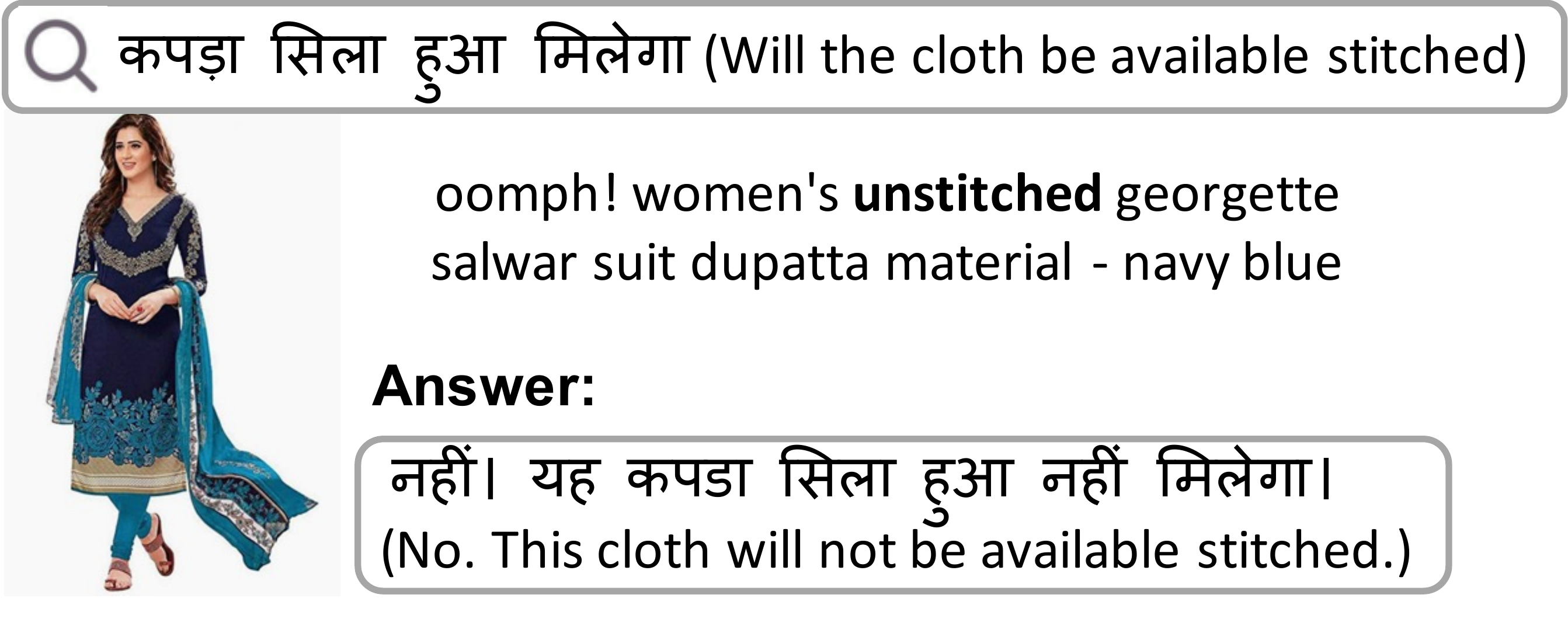}
    \caption{\small Cross-lingual PQA: The user asks questions about a product in their language (such as Hindi), then the system searches for product information in English and generates an answer in the same language as the question.
    }
    \label{fig:intro}
\end{figure}

To address this, we present xPQA, the first large-scale dataset for cross-lingual PQA enabling non-English questions to be answered from English content. 
Most comprehensive product information is usually available in a majority language such as English. Therefore, searching for relevant information in English often has a better chance of finding an answer.\footnote{Prior work~\cite{asai2021xor} also shows the effectiveness of using English data as a knowledge source for cross-lingual QA. It is nevertheless helpful to also support searching in all languages, which we leave for future work.}
This paper explores how to effectively \emph{train systems that retrieve information from English and generate answers in the question language to allow users to ask questions in any language.} Fig~\ref{fig:intro} shows an example.

Most existing multilingual QA datasets are created by translating English questions, introducing translation artifacts and discrepencies from native speakers' real information-seeking behaviors~\cite{clark2020tydi}. Instead, we collect questions from the original market places as written by native speakers, hire bilingual annotators to check the relevant product information and write the final answers in their target languages. This eliminates the need for translations and ensures that the information-seeking behaviors of native speakers are accurately represented.


Based on the collected dataset, we report baseline results on two subtasks: (a) candidate ranking, which selects the best English candidate that contains the information to answer the non-English question; (b) answer generation, which generates a natural-sounding non-English answer to present to the user based on the selected English candidate. 
We find that applying a cross-lingual ranker trained on a Wikipedia-based QA dataset generalizes poorly to the product domain. 
The performance is even worse than training a multilingual ranker on the English in-domain data, suggesting that domain transferability is even more crucial than language transferability. The translation-based approach is the most effective for candidate ranking while the multilingual-finetuning works the best for answer generation.
Nonetheless, on both tasks, there is a substantial gap between the English-based and cross-lingual performances. In the following, we first elaborate on the problem formulation for the cross-lingual PQA task (\cref{sec:problem_formulation}), then explain the xPQA data collection process (\cref{sec:data_collect}), and present experiment results (\cref{sec:results}) and conclusions
(\cref{sec:conclusion}).

\section{Problem Formulation}
\label{sec:problem_formulation}
\paragraph{Task}
There are two important tasks for a cross-lingual PQA system: \emph{candidate ranking} and \emph{answer generation}. 
In candidate ranking, given a question in a target language and a list of candidates in English, the ranker 
predicts a relevance score for every candidate and selects the top one.
Candidate ranking is necessary because a given product webpage may contain hundreds of information pieces about the product, so as a practical matter we select the top candidate to use in generation. 
After getting the top candidate, an answer generator takes it as input together with the question and produces an answer in the question language. This step is crucial in order to deploy a user-friendly PQA system since the candidate is neither in the user language nor written specifically to answer the question.
\paragraph{Scenario}
We consider two scenarios for both tasks: \emph{zero-shot} and \emph{fine-tuned}. 
\emph{Zero-shot} assumes that we do not have any labeled data and must rely on transfer learning from the English-based PQA dataset~\footnote{Another option is transfer learning from cross-lingual datasets in other domains, as we evaluate later.}. 
\emph{Fine-tuned} assumes that we can further finetune models on a limited number of cross-lingual PQA annotations. 
Both are realistic scenarios as annotations are usually more abundant in English than in other languages~\cite{shen2023neural}. In our experiments, we use ePQA as the English-based PQA dataset, which is an extension of the dataset in \citet{shen2022product} with coverage and quality improvements. Details are in Appendix~\ref{app:diff_hetqa}.

\section{xPQA Dataset Collection}
\label{sec:data_collect}
To train and evaluate our two tasks, the xPQA dataset contains annotations for (1)~question-candidate relevance to label whether every candidate is relevant to the question or not, and (2)~answers where a natural-sounding answer is manually written if the candidate contains enough information to address the question. The collection process follows the steps below:
\begin{table}
\centering
\small
\begin{tabular}{llll}
\toprule
\textbf{Language} & \textbf{Branch} & \textbf{Script} & \textbf{Market}\\\midrule
German (DE) & Germanic & Latin& Germany\\
Italian (IT) & Romance & Latin&Italy\\
French (FR) & Romance & Latin&France\\
Spanish (ES) & Romance & Latin&Spain\\
Portuguese (PT) &  Romance & Latin&Brazil\\
Polish (PL) & Balto-Slavic & Latin&Poland\\
Arabic (AR) & Semitic & Arabic& SA\\
Hindi (HI) & Indo-Aryan & Devanagari & India\\
Tamil (TA) & Dravidian & Tamil & India\\
Chinese (ZH) &  Sinitic & Chinese& China\\
Japanese (JA) & Japonic & Kanji;Kana & Japan\\
Korean (KO) &  Han &Hangul& US\\
\bottomrule
\end{tabular}
\caption{\label{tab:lang_stats}\small
Languages in the xPQA dataset.
}
\end{table}

\paragraph{1. Question Collection}
For our question set, we crawl publicly-available community questions from Amazon.com product pages in 11 markets, obtaining questions in 12 different languages.
For each language, we choose the corresponding market, then sample 2,500 unique questions. From these sampled questions, we select 1500 questions for each language that are manually verified by our annotators as being in the target language, information seeking, and containing no offensive content. 

\paragraph{2. Candidate Collection}
For every valid question, we link its corresponding product page in the US market (except for Hindi and Tamil which directly use the India market) and extract all English candidates from product information sources (details in Appendix~\ref{app:cand_process}). Then, we translate every question into English with AWS translate,\footnote{\url{https://aws.amazon.com/translate/}} feed the translated question into an English-based ranker~\footnote{An ELECTRA~\cite{clark2020electra}-based binary classifier model pretrained on large amounts of pseudo labels plus human annotations optimized for the English ranking task.} and obtain top-5 candidates from its candidate set. 

\begin{figure*}[t]
    \centering
    \includegraphics[width=1.95\columnwidth]{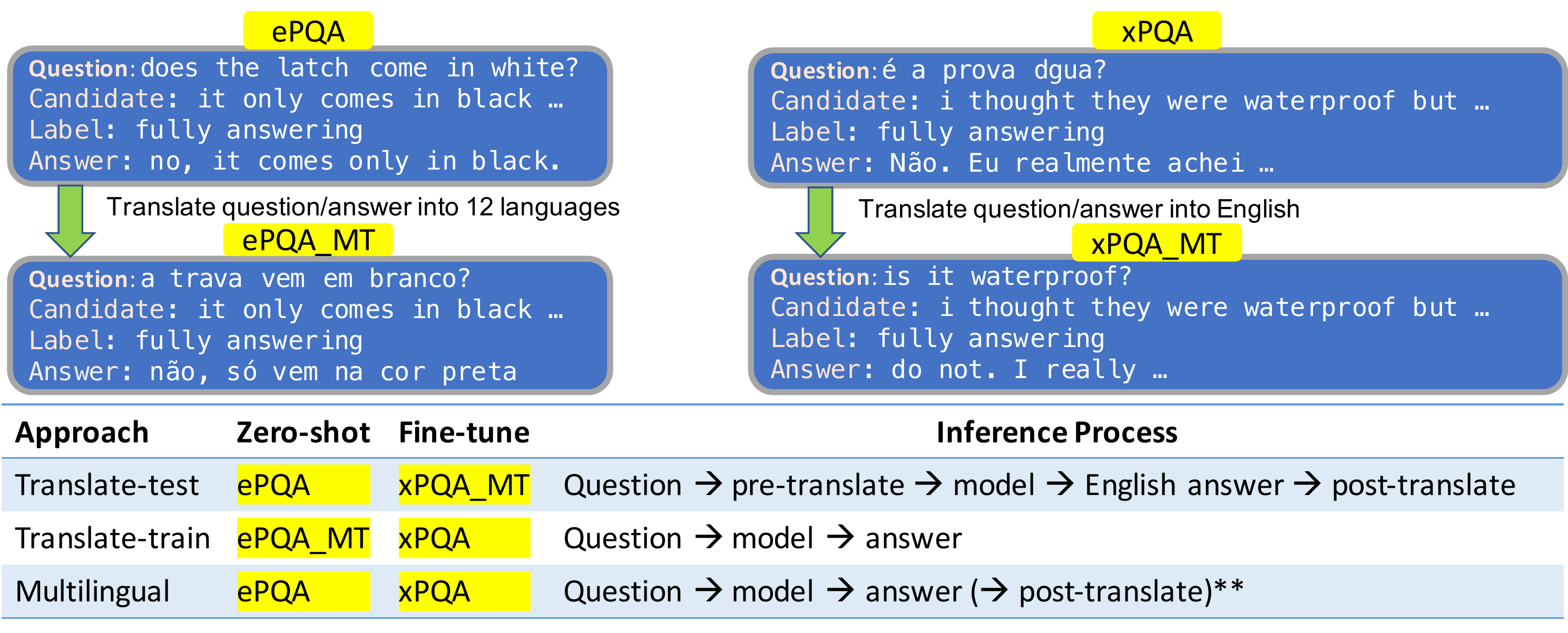}
    \caption{\small \textbf{Summary of experimented approaches}. The ePQA\_MT (and xPQA\_MT) set is the translated version of ePQA (and xPQA) into all non-English languages (and English). **indicates that post-translate is only required for the zero-shot model.
    }
    \label{fig:model}
\end{figure*}

\paragraph{3. Relevance Annotation} The top-5 English candidates and the non-English original questions are passed to annotators to judge their relevance.
Each candidate is marked with one of three labels: ``fully answering'' (contains enough information to address the question), ``partially answering'' (contains useful information to partially address the question), and ``irrelevant'' (does not provide any helpful information). Guidelines are available in Appendix~\ref{app:rel_annot}.
\paragraph{4. Answer Search} To increase the answer coverage,  questions for which none of the top-5 candidates are marked as ``fully answering'' are given to annotators who are asked to actively search for the answer on the Amazon product page. If they find candidates fully answering the question, these are included with the label ``fully answering''.
\paragraph{5. Answer Generation}
For candidates marked as ``fully answering'', annotators are then asked to write natural, direct answers based on them.

All annotators are bilingual, hired through the centific platform~\footnote{\url{https://www.centific.com/}}. 
The constructed xPQA dataset is split into 1000/400/100 questions as the test/train/dev sets for each language. Table~\ref{tab:lang_stats} shows all languages included in the xPQA dataset. The detailed annotation process, payment, and statistics are explained in Appendix~\ref{app:data_collection}. 

\section{Approaches}
For each task, we experiment with three types of baseline approaches: \textbf{translate-test}, \textbf{translate-train}, and \textbf{multilingual}~\cite{hu2020xtreme}. Fig~\ref{fig:model} provides a summary of these approaches. 

\paragraph{Translate-test} 
The essential idea here is to rely exclusively on English-centric models and datasets. 
In the zero-shot scenario, models are trained on the ePQA dataset. 
In the fine-tuned scenario, we must translate questions and answers in the xPQA dataset into English as this is an English-centric model. 
This translated dataset, termed xPQA\_MT is used to further fine-tune the zero-shot models. 
At runtime, we use an external machine translation model to translate the question into English and apply the ranker to select the best candidate. 
Afterwards, an English-based generator produces an answer in English, which is then post-translated to the target language. 
\textbf{Translate-test} is a common approach in industry as it uses well-trained English-based models and off-the-shelf translation tools without further modifications. However, such a pipelined process introduces runtime latency and can lead to error propagation if translation quality is not perfect.

\paragraph{Translate-train} In contrast to the above, here we apply all translation processes in training, or offline, so that no additional latency is added at runtime.
In the zero-shot scenario, we machine-translate all questions and answers in the ePQA dataset into each of the 12 languages we consider. The resulting dataset, termed ePQA\_MT, is used to train a multilingual model. 
In the fine-tuned scenario, we further finetune the model on the xPQA dataset. 
As the model is defined to be multilingual, it can directly take input questions in their original languages and output answers in the target languages without any translation process.

\paragraph{Multilingual} Finally, this approach is similar to the translate-train one in that both use multilingual models rather than an English-only model, but the difference is that the multilingual approach requires no translations at training time. 
In the zero-shot scenario, it trains a multilingual pretrained model directly on the English-only ePQA dataset and relies only on its own pretrained multilingual knowledge to adapt to other languages. 
In the fine-tuned scenario, we further fine-tune the model on the xPQA dataset. 
Note that this approach still requires runtime post-translation of the generated English answer into the target language. This is because we find that multilingual models can only generate English answers when trained only on English datasets. Although special decoding constraints could be use to restrict output vocabulary to that of the target language, zero-shot multilingual adaptation in generation tasks is still an open challenge~\cite{chen2022mtg,zhang2022mdia}.

It is worth mentioning that \emph{the three types of approaches can be combined}.
For example, we could follow the \textbf{translate-train} approach to train the candidate ranker and follow the \textbf{multilingual} approach to train the answer generator. Details of the model implementation are in Appendix~\ref{app:exp}.


\begin{table*}
\centering
\small
\begin{tabular}{l|cccccccccccc|c}
\toprule
Model&DE&IT&FR&ES&PT&PL&AR&HI&TA&ZH&JA&KO&AVG\\
\midrule
\multicolumn{14}{c}{\textbf{Zero-shot Scenario}}\\
\hline
Translate-test & \textbf{48.7}&\textbf{48.6}&\textbf{59.7}&\textbf{63.8}&56.9&\textbf{63.6}&\textbf{49.2}&\textbf{60.2}&\textbf{44.6}&\textbf{56.1}&\textbf{50.7}&48.9&\textbf{54.2}\\
Multilingual &48.4&46.2&59.1&59.8&55.5&60.0&45.1&42.7&40.4&53.0&45.0&45.4&50.1 \\
Translate-train &47.7&47.8&57.4&60.8&\textbf{57.0}&58.7&48.7&50.9&44.1&55.8&47.8&\textbf{49.8}&52.2 \\
\midrule
\multicolumn{14}{c}{\textbf{Fine-tuned Scenario}}\\
\hline
Translate-test & 51.7&\textbf{55.1}&\textbf{64.8}&\textbf{66.8}&\textbf{64.0}&68.0&\textbf{57.3}&\textbf{68.4}&50.0&61.9&\textbf{57.9}&60.2&\textbf{60.5}\\
Multilingual &\textbf{52.7}&53.5&\textbf{64.8}&65.7&63.5&70.6&54.7&67.6&49.0&60.3&51.6&57.8&59.3 \\
Translate-train &52.1&54.0&63.4&67.1&62.1&\textbf{71.6}&55.1&67.4&\textbf{51.3}&\textbf{64.2}&54.7&\textbf{60.6}&60.3 \\
\bottomrule
\end{tabular}
\caption{\label{tab:ranking_results}\small
P@1 of candidate ranking for each language and the averaged score (AVG) on answerable questions in xPQA testset.
}
\end{table*}

\section{Experiment}
\label{sec:exp}
\subsection{Evaluation}
Although many QA works report end-to-end performances, we chose not to report them because (1) Most product questions, as well as the information sources such as reviews and customer answers, are subjective. The correctness of answers depends on the specific candidates for which there is no universal ground truth~\cite{mcauley2016addressing}; (2) Only providing answers grounded on references is a critical requirement for an online PQA deployment. 
When candidate ranking fails to provide suitable candidates in the first stage, even if the answer generator manages to make a good guess,\footnote{As the answer depends on the information of the specific product, the chance of guessing the correct answer without proper candidates is close to random and fully unreliable. } it is still considered a failure. Therefore, end-to-end evaluations are not suitable and the evaluation of answer generation has to be candidate-dependent.

We evaluate the ranker with Precision of the top-1 candidate, P@1, as the generated answer is based on the top-1 candidate. To remove the effects of language-specific answering ratios, we report P@1 scores only on the answerable questions where at least one candidate is marked as ``fully answering''. The generator is evaluated with the sacreBLEU score~\footnote{\url{https://github.com/mjpost/sacrebleu}}. The generations are produced and evaluated only from candidates marked as ``fully answering'' since otherwise, the ground truth is undefined. 

\subsection{Main Results}
\label{sec:results}
\paragraph{Task 1: Candidate Ranking}
Table~\ref{tab:ranking_results} shows P@1 of different candidate ranking approaches and their average scores.
\textbf{Translate-test} performs the best, and its advantage is particularly prominent in the zero-shot scenario. 
In the fine-tuned scenario, however, the other two approaches can also perform similarly. The \textbf{translate-train} approach outperforms the \textbf{multilingual} approach mainly for languages that do not use Latin scripts. Even for low-resource languages, such as Tamil whose translation quality is far from satisfactory, translating the training corpus still helps the multilingual model adapt to the target language. 
This implies \emph{existing pre-trained multilingual models are already good at adapting to new languages with Latin scripts. Translating the training corpus is mainly helpful to adapt the model into new scripts}~\cite{lauscher-etal-2020-zero}. Fine-tuning an English BERT on the ePQA training set leads to a P@1 of 70.7\% on the monolingual English test set, which is significantly higher than all other languages except Polish, suggesting scope for substantial improvement.\footnote{Note that \emph{this does not mean the system works better for Polish than English}. As the question set for each language is distinct and not comparable, it could be simply that the sampled Polish questions are easier than English ones.}

\begin{table*}
\centering
\small
\begin{tabular}{l|cccccccccccc|c}
\toprule
Model&DE&IT&FR&ES&PT&PL&AR&HI&TA&ZH&JA&KO&AVG\\
\midrule
\multicolumn{14}{c}{\textbf{Zero-shot Scenario}}\\
\hline
Translate-test & 7.0&17.1&14.3&11.5&19.4&11.7&\textbf{18.5}&8.9&\textbf{5.1}&\textbf{19.8}&\textbf{12.9}&\textbf{8.5}&12.9\\
Multilingual &6.0&14.2&11.6&10.1&18.3&9.9&16.3&7.0&4.8&17.8&11.7&5.9&11.1 \\
Translate-train &\textbf{16.9}&\textbf{17.1}&\textbf{20.5}&\textbf{14.1}&\textbf{19.5}&\textbf{18.8}&15.9&\textbf{15.8}&4.4&16.6&12.8&7.4&\textbf{15.0} \\\midrule
\multicolumn{14}{c}{\textbf{Fine-tuned Scenario}}\\\hline
Translate-test & 8.9&25.3&15.4&14.2&21.0&16.6&\textbf{17.3}&16.3&7.1&21.7&12.3&8.8&15.4\\
Multilingual &27.2&27.1&22.5&31.3&20.0&32.3&13.4&26.0&16.7&26.2&31.6&44.0&26.5 \\
Translate-train &\textbf{32.9}&\textbf{31.6}&\textbf{26.6}&\textbf{36.6}&\textbf{24.4}&\textbf{40.1}&16.0&\textbf{28.5}&\textbf{18.5}&\textbf{30.3}&\textbf{33.7}&\textbf{51.6}&\textbf{30.9} \\
\bottomrule
\end{tabular}
\caption{\label{tab:generation_results}
\small BLEU score of answer generation for each language and the averaged score (AVG) on the xPQA test set.
}
\end{table*}

\begin{figure}[t]
    \centering
    \includegraphics[width=\columnwidth]{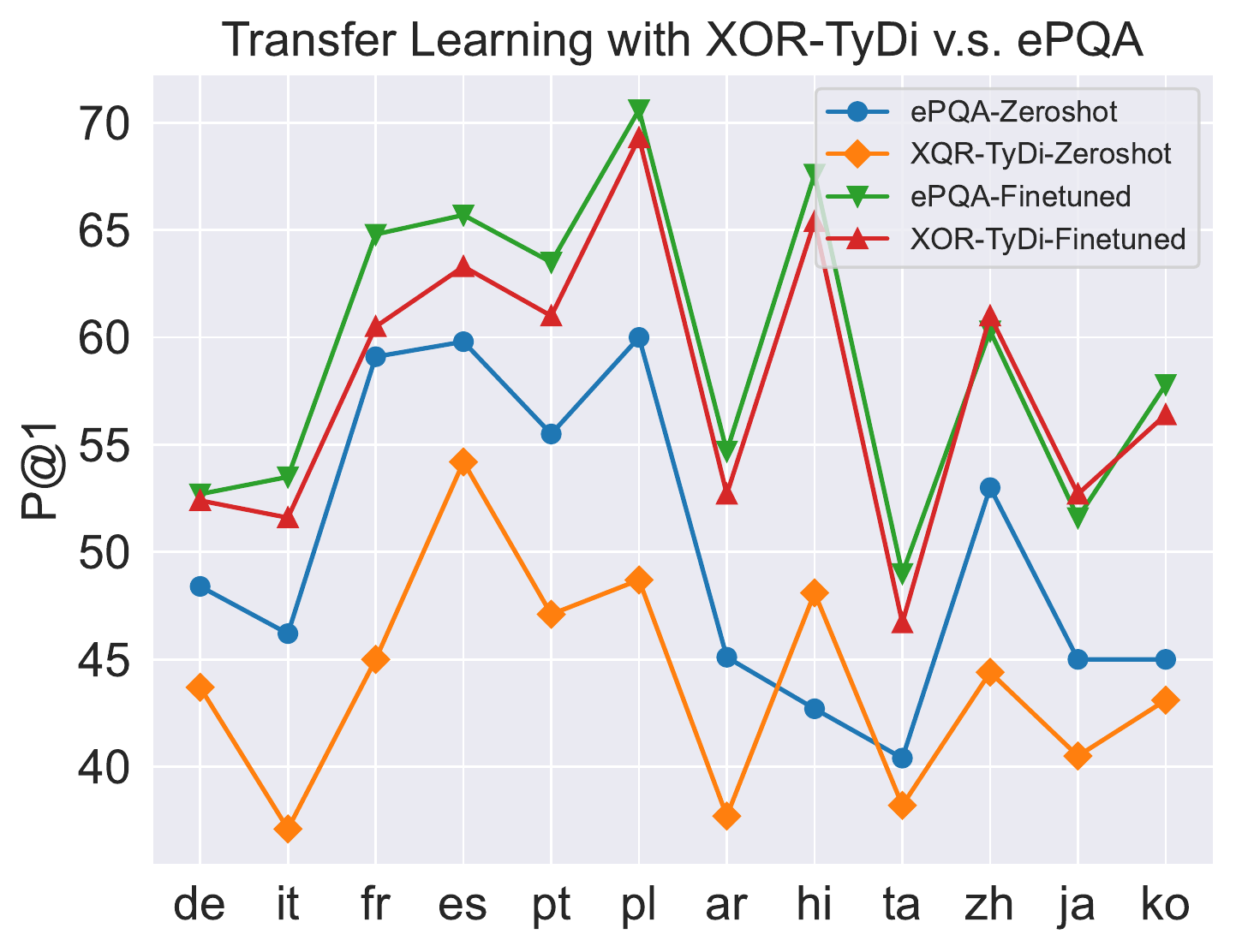}
    \caption{\small Comparison of transfer learning datasets. The English-only in-domain ePQA data is more useful than the cross-lingual out-of-domain XOR-Tydi dataset.
    }
    \label{fig:transfer_ablation}
\end{figure}

\paragraph{Task 2: Answer Generation}
Table~\ref{tab:generation_results} shows the BLEU score of different answer generation approaches and their average scores. In the zero-shot scenario, the \textbf{translate-test} approach often performs the best on languages with non-Latin scripts and the \textbf{translate-train} approach performs the best on languages with Latin scripts. The \textbf{translate-train} approach outperforms the \textbf{multilingual} approach with a few exceptions. Interestingly, all the exceptions happen in languages using non-Latin scripts, which contradicts the findings in candidate ranking. 
We hypothesize that the used pre-trained multilingual model is better at understanding non-Latin scripts than actually generating them because generating the text requires more advanced knowledge of grammar, which cannot be easily distilled from imperfect machine translators~\cite{adelani2022few}. Fine-tuning models on the xPQA training data leads to big improvements across approaches, especially for \textbf{multilingual} and \textbf{translate-train} which do not rely on machine translators at runtime. The \textbf{translate-test} approach, due to the error propagation from two machine translation steps, significantly underperforms the other two. Fine-tuning an English T5 model on the ePQA training set leads to a BLEU score of 49.7\%; although BLEU scores are related to language-specific tokenizers and questions, we believe this consistent gap implies large opportunities for improvement.

\subsection{Analysis}
\paragraph{Domain vs Language Transferability}
There are cross-lingual QA datasets in other domains. When building a system for xPQA, is it better to use an English-only in-domain QA dataset or a cross-lingual out-of-domain QA dataset?
To answer this question, we train a new multilingual ranker on the XOR-TyDi dataset~\cite{asai-etal-2021-xor}, which is a representative cross-lingual QA dataset with real questions from the Wikipedia domain. 
We treat the gold passage containing the correct answer span as positive and randomly sample 5 other passages as negative. The comparison with our existing \textbf{multilingual} approach trained on the ePQA dataset is shown in Figure~\ref{fig:transfer_ablation}. We can see that fine-tuning models on the ePQA dataset leads to significantly better performance on all languages with few exceptions, suggesting \emph{domain
differences are even more crucial than language differences
for the candidate ranking task in xPQA}. 
It is necessary to collect in-domain annotations for good performance.
\begin{figure}[t]
    \centering
    \includegraphics[width=\columnwidth]{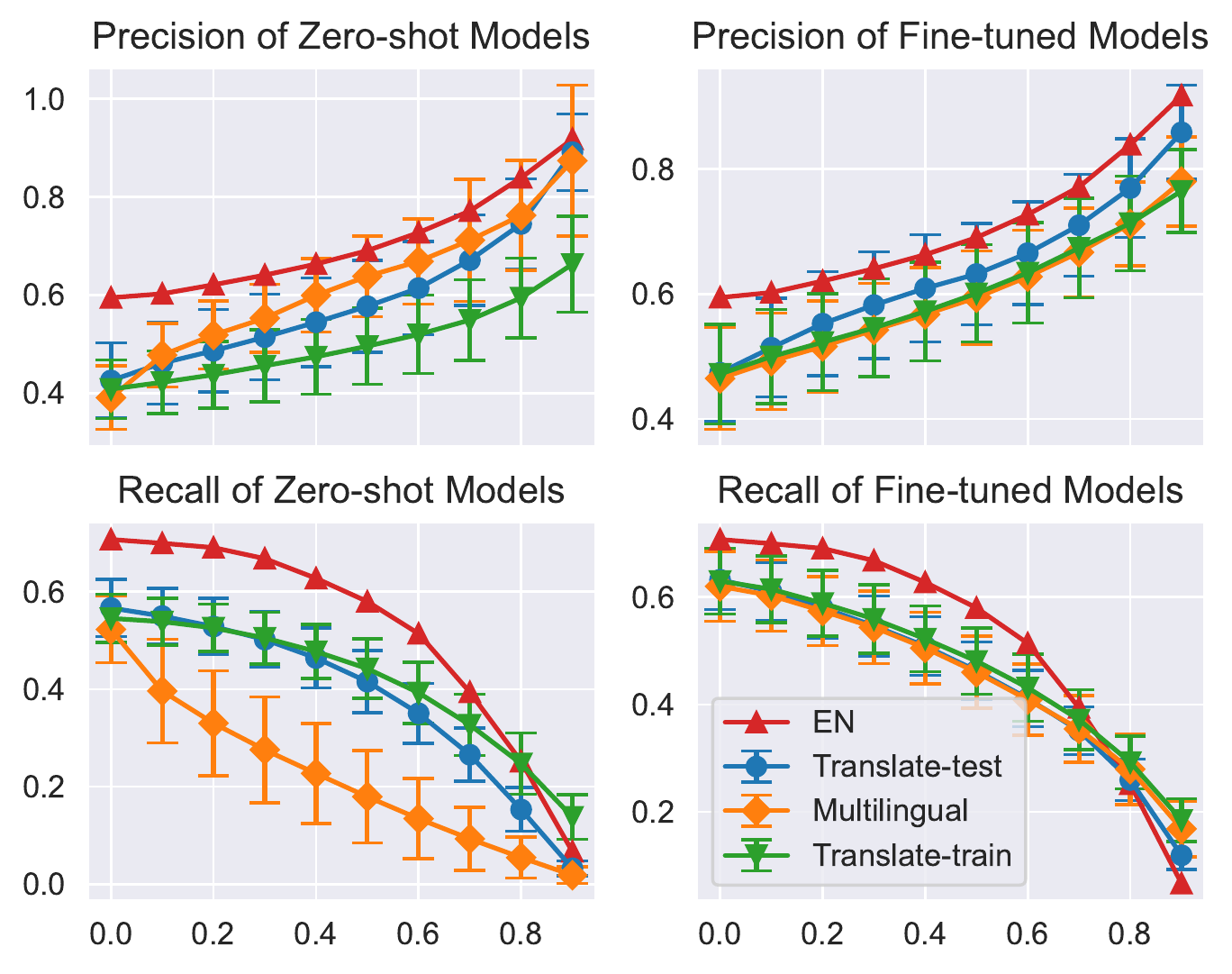}
    \caption{\small Precision and recall with varying thresholds. The red line is for the English model and the other lines are for the average score of the cross-lingual model. Vertical bars are the standard deviations across all languages.
    }
    \label{fig:prec_rec}
\end{figure}

\paragraph{Answerability Prediction}
As the amount of information differs among products, it is very likely that many questions are not answerable with existing candidates and the model should not attempt to answer given the available information. A common practice is to use the model score as a predictor for the answerability confidence. To see how effective this is, we visualize the change of precision and recall with varying model score thresholds in Figure~\ref{fig:prec_rec}. We can see that in the zero-shot scenario, there is a larger performance variance across languages, especially for the \textbf{multilingual} approach which solely relies on the knowledge from the pre-trained model. The \textbf{multilingual} approach is also more sensitive to the threshold and its recall drops much faster than the other two approaches. 
Fine-tuning the xQA training data reduces the gaps between the three approaches. 
The English model, as expected, consistently performs better, especially in the low-confidence region.
\begin{table}
\centering
\small
\begin{tabular}{ccccc}
\toprule
\textbf{Language} & \textbf{P@1} & \textbf{AUPC} & \textbf{MAP} & \textbf{MRR}\\\midrule
DE (MT) & 48.7 & 61.8 & 64.4&66.4\\
DE (HT) & 48.8 (\textbf{$\uparrow$0.1}) & 61.9& 64.5&66.5\\
TA (MT) & 44.6 & 59.7 &60.0 &62.3\\
TA (HT) & 54.2 (\textbf{$\uparrow$9.6}) & 69.8&66.3&69.1\\
\bottomrule
\end{tabular}
\caption{\label{tab:ht_mt}\small
Comparison of translate-test approach (candidate ranking) using Machine (MT)/human translation (HT) .
}
\end{table}
\paragraph{Effects of Translation Quality}
To investigate the effects of the translation quality in the \textbf{translate-test} approach, we select German and Tamil as two languages with very different translation qualities and obtain manual translations of their questions. Comparisons to machine-translated questions are shown in Table~\ref{tab:ht_mt}. Apart from P@1, we also show AUPC (Area Under Perturbation Curve), MAP (Mean Average Precision) and MRR (Mean Reciprocal Rank) scores. We can see that the improvement from using human translations is negligible in German but substantial in Tamil. Even with human translations, we can still see a big gap between performances on English monolingual (70.7\%) and xPQA test sets (48.8\% and 54.2\%), suggesting that \emph{question-shape shifts can be even a bigger challenge than language shifts} for the candidate ranking task. The problem of language shifts might be crucial only for low-resource languages without decent MT systems such as Tamil.

\begin{table}
\centering
\small
\begin{tabular}{cccc}
\toprule
\textbf{Pre-Translate} & \textbf{Rank} & \textbf{Generate} & \textbf{Post-Translate}\\\midrule
74.1ms & 21.3ms & 532.4ms& 91.3ms\\
\bottomrule
\end{tabular}
\caption{\label{tab:latency}\small
Latency of each component. Generating and translating cost much more time than ranking.
}
\end{table}
\paragraph{Runtime Latency}
Table~\ref{tab:latency} shows the runtime latency of every component tested in one AWS P3.16 instance. We feed questions in all languages one by one to simulate an online environment. As seen, the candidate ranker is fast and the computation over multiple candidates can be easily parallelized. The pre/post-translate costs more time, but the main bottleneck is the answer generation step, which is $25\times$ slower than the ranking. This is clearly more than the latency budget of most online applications and can be the focus of future research. Potential improvements could be in non-autoregressive decoding, efficient attention, or distillation into a smaller model~\cite{tang2021ast,tang2022ast,li2022dq}. 
\section{Conclusion}
\label{sec:conclusion}
This paper presents xPQA, a dataset for cross-lingual PQA supporting non-English questions to be answered from English product content. 
We report baseline results and findings for three approaches: translate-test, multilingual, and translate-train. Experiments show that the translate-test approach performs the best for the candidate ranking task while the translate-train approach performs the best for the answer generation task. However, there remains significant room for improvement relative to an English-based monolingual PQA system. 
We hope that future research can benefit from our work to improve cross-lingual PQA systems.

\section*{Limitations}
While the xPQA dataset is created to be as close to the real-world scenario as possible, it has two major drawbacks. Firstly, the candidate set in the dataset does not include the full candidates for a given product because annotating all candidates is prohibitively expensive. The subjectivity of product questions and candidates also makes it hard to get ground-truth short-span answers, which prevents a straightforward end-to-end evaluation over the full candidate set. A potential fix is to run human evaluations on the top-1 candidate over the full candidate set from each model, but it'd be costly to do so. A more realistic solution is to have an online evaluation for the best model only, which we leave for future work. Secondly, the answer annotation is based only on a single candidate because handling information from multiple candidates requires careful instructions on conflicting information and summarization skills. This might limit the model in answering complex questions that require inference over multiple candidates. However, we find this case to be very rare in real customer questions. Furthermore, as we do not summarize multiple candidates, the returned answer can be biased toward the opinion of a single customer. 
Our evaluation also has potential limitations in that (1) We did not extensively evaluate the quality of generated answers with manual annotation. It is known that BLEU scores might not correlate well with human evaluations on generation tasks, and they can be misleading in certain cases; (2) We only compared major types of baseline algorithms and did not explore the effects of leveraging existing larger, more powerful pre-trained language models such as mT0~\cite{muennighoff2022crosslingual} and Flan-T5~\cite{chung2022scaling}. Conclusions might change if we hire annotators to perform more human evaluations or change the model architecture.
\section*{Ethics Statement}
E-commerce has been increasingly popular these years. Nonetheless, a big amount of people cannot benefit much from it because most E-commerce websites only support a few major languages. Deploying an xPQA system can have a broad impact across a wide range of non-English speakers to assist them in their shopping experience. With a well-developed xPQA system, we only need to maintain comprehensive product information in one majority language, but allow non-English speakers easily get access to the product information. This can significantly reduce the maintenance cost and benefit the democratization of AI. Nevertheless, there are two major caveats before deploying a safe, reliable xPQA system: (1) The answer generator needs to be fairly evaluated by humans in terms of faithfulness. While answer generation can greatly improve user-friendliness, it also brings potential risks of providing false information; (2) The users should be well noticed that the provided answer is drawn from the opinion of a single customer or other sources. It cannot reflect the opinion of the vendor, or seller nor imply any trend from the public.

\bibliography{anthology,custom}
\bibliographystyle{acl_natbib}

\appendix

\section{Difference with Previous Datasets}
\label{app:diff_hetqa}
\paragraph{Product Question Answering} Product question answering (PQA) differs from general-knowledge QAs in that questions often seek subjective opinions on specific products, so earlier research usually treated it as an opinion mining problem~\cite{moghaddam2011aqa,yu2012answering}. Recent advances in neural networks propagated the use of dense retrieval and generation models to provide direct answers. Many relevant datasets are curated to facilitate this study~\cite{chen2019driven,xu2020data,gao2021meaningful,deng2022toward,shen2022semipqa,shen2022low}. However, they are either based on simulated questions, or community question-answers where the answers are noisy and have no direct connection with product information candidates~\cite{lai2018simple,xu2019review,barlacchi-etal-2022-focusqa}. The only exception is \citet{shen2022product} where exact annotations are provided for both candidate relevance and answer generation, but it focuses only on one product category and the annotation quality is not good enough. Specifically, we sample about 2000 question-candidate pairs then perform an in-house annotation and find around 20\% of the annotations are incorrect. As a result, we construct the ePQA dataset with the following main differences from the dataset in \citet{shen2022product}: (1) It has higher annotation quality with rounds of verifications. In our in-house annotation, the error rate is less than $5\%$; (2) It does not restrict the product categories, while the original dataset focuses only on the toys and games products; (3) It defines finer-grained 3-class labels for each candidate, while the original dataset contains only binary labels; (4) Every candidate is checked with its context (surrounding sentences) to make sure the label is correct.

To the best of our knowledge, all existing PQA datasets are monolingual and questions are usually in high-resource languages such as English or Chinese, which leads to our motivation of building a cross-lingual PQA dataset.

\paragraph{Cross-Lingual Question Answering}
Recently, many non-English question answering (QA) datasets in the general Wikipedia domain have been proposed~\cite{lewis-etal-2020-mlqa,artetxe-etal-2020-cross,clark-etal-2020-tydi,hardalov-etal-2020-exams}. 
Several datasets focus on the \emph{open-retrieval (open-domain)} setting, where a gold document or paragraph is not pre-given and a system needs to search documents to answer questions~\cite{liu-etal-2019-xqa,asai-etal-2021-xor,longpre-etal-2021-mkqa}. 
Importantly, all of those prior datasets are created based on Wikipedia or school exams, and there is no prior work on cross-lingual product QA.

\begin{figure*}[t]
    \centering
    \includegraphics[width=2\columnwidth]{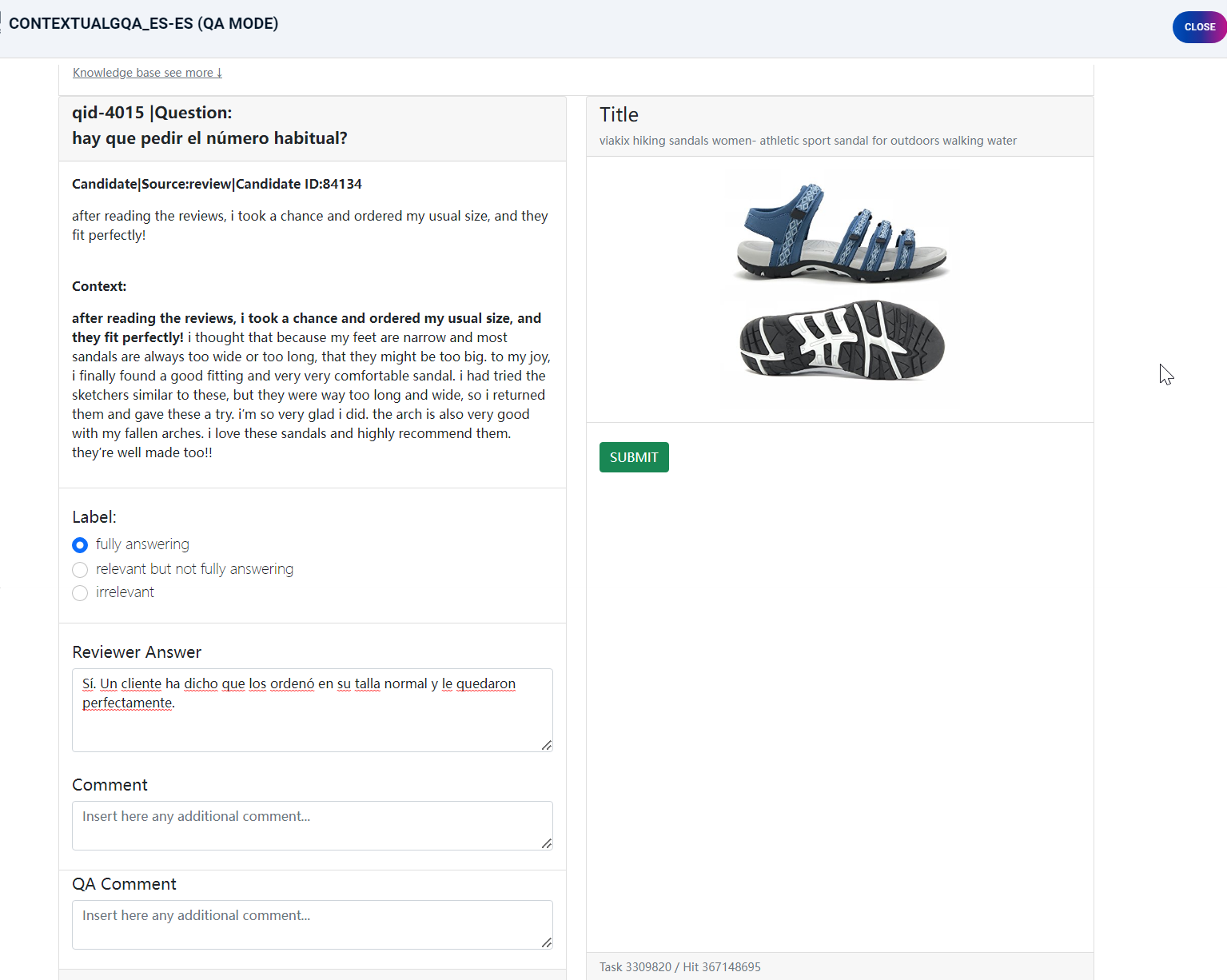}
    \caption{\small UI of the annotation task. Annotators will be shown a question in one of the 13 languages we considered and a candidate extracted from product information. Annotators can also see the title, and picture of the product, as well as context (surrounding sentences of the candidate with the actual candidate being highlighted), to provide a more accurate annotation.
    }
    \label{fig:anno_ui}
\end{figure*}

Notably, \emph{ePQA} contains 131,52/1,000/2,000 questions in the train/dev/test sets respectively, which is significantly larger than xPQA (as in realistic scenarios). 
It can be used to analyze the performance gap between mono-lingual PQA and cross-lingual PQA. 

\section{Dataset Collection}
\label{app:data_collection}
\subsection{Question Collection}
\label{app:question_collect}
In the question collection phase, questions are kept if they fulfill the following criteria: (1) It is identified as the target language through Amazon Comprehend~\footnote{\url{https://aws.amazon.com/comprehend/}}; (2) It contains no URL links; (3) It contains at most one question mark so as to avoid multiple questions; (4) It contains at least 3 words and less than 20 words; (5) Its corresponding product is also available in the US market.~\footnote{This is to ensure we can get English candidates for these products. It does not apply to Hindi and Tamil because the official language in the India market is already English.}

\subsection{Candidate Processing}
\label{app:cand_process}
Our candidates come from 6 information sources: (1) product title, (2) semi-structured attributes, (3) product bullet points, (4) product description, (5) community answers (excluding the answer that directly replies to the question); (6) user reviews. Every product title and attribute is treated as a single candidate. 
For the other product information, we split them into sentences and treat each sentence as the candidate. For candidates from community answers, We further concatenate them with the corresponding community questions to provide more context. All candidates are lower cases and emojis are removed. Numbers from the semi-structured attributes are further normalized to keep at most 2 decimals.

\begin{figure*}[t]
    \centering
    \includegraphics[width=2\columnwidth]{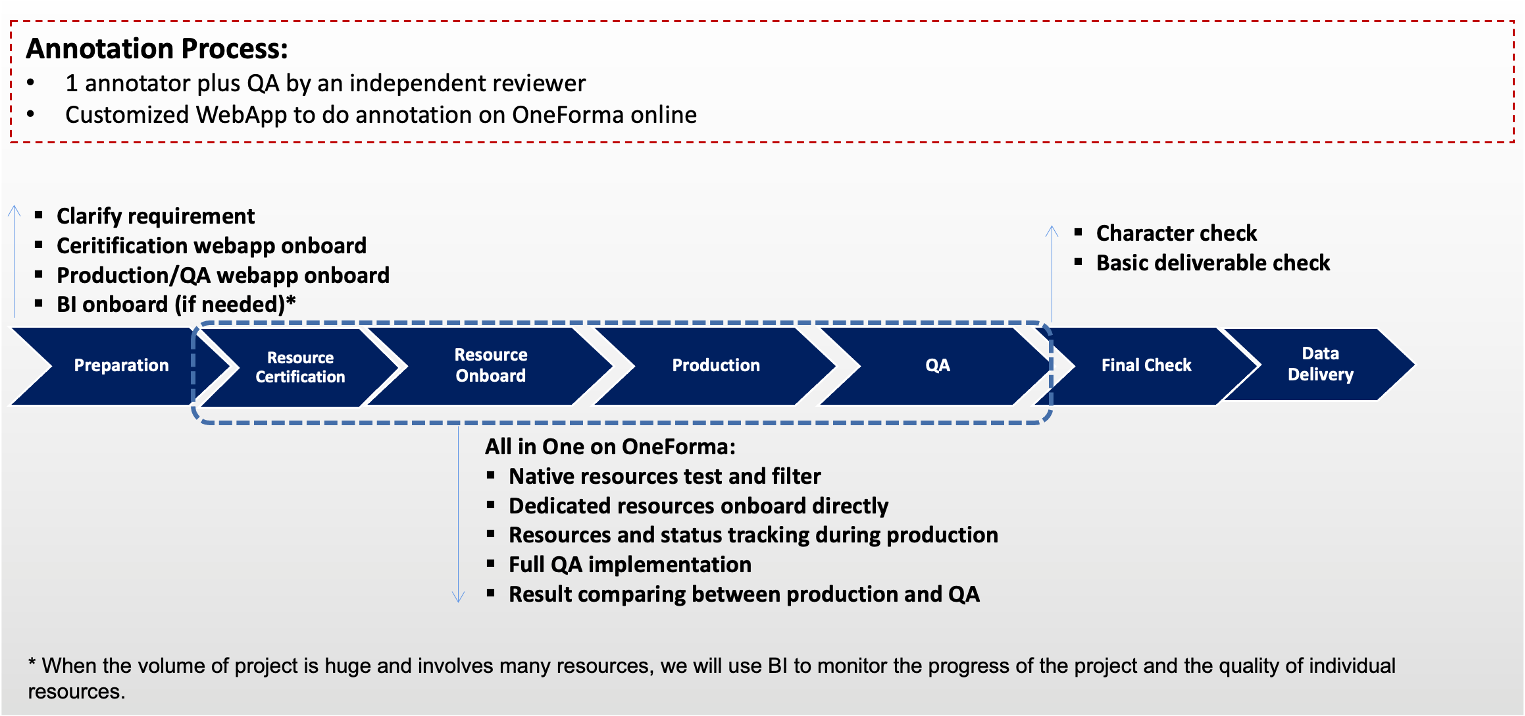}
    \caption{\small Annotation process and quality control of the task.
    }
    \label{fig:anno_process}
\end{figure*}

\subsection{Relevance Annotation}
\label{app:rel_annot}
Each candidate is marked with one of three labels: ``fully answering'' (it contains enough information to address the question), ``partially answering'' (it contains useful information to partially address the question), and ``irrelevant'' (it's not useful in answering the question at all). To make sure the candidate is properly understood, we also provide its context (surrounding sentences) to the annotators. The exact definitions  for the three labels and guidelines used are:
\begin{itemize}
    \item Fully answering. Meaning that the response contains clear information to tell about the answer. It can take some inference step to get the answer, but it must contain enough information to help come to the answer.
    \item Partially answering (relevant but not fully answering). Meaning that the response contains useful information that help one understand more, and narrow down the range of the answer, yet not enough to get the exact answer from it.
    \item Irrelevant. Meaning that the response does not provide useful relevant information at all, and a customer will not get anything new about their question after reading it.
\end{itemize}
Note that in this step, annotators do NOT need to consider factual correctness. For the question “what color is it?”, it does not matter if the response is saying it is blue or red. Annotators should focus on the content only but not the factual correctness.
Besides, even if it contains other extra information or the response is not natural, as long as the proper information is included, then it is considered as fully answering.

Specifically, Fully answering means the response contains enough information to let one draw the answer. The criteria of fully answering should NOT be overly strict. Annotators can be lenient with the word choice, as long as the response conveys the proper meaning. For example:

Question: is it an awesome gift for my girl friend?
Response: it is a nice valentine gift for your partner.

In this case, the difference between “awesome” and “nice” is not relevant, as the response is either way saying that it is a good gift for your girl friend or partner, and thereby should be judged as “fully answering”.

Another example:

Question: is it comfortable to sleep on for a 6” tall man?
Response: It is comfortable to lie down for tall people.

Annotators should not be overly strict about whether 6” can be considered as “tall” and whether “lie down” is equivalent to “sleep on”, etc. Based on common sense, if the immediate impression after reading the response provides the needed information, one should NOT overthink other ways of interpreting this response.

\begin{table*}
\centering
\small
\begin{tabular}{l|cccccccccccc|c}
\toprule
Task&DE&IT&FR&ES&PT&PL&AR&HI&TA&ZH&JA&KO&EN\\
\midrule
\multicolumn{14}{c}{\textbf{Zero-shot Scenario}}\\
\hline
Relevance Annotate & 0.30&0.22&0.30&0.22&0.22&0.24&0.22&0.13&0.19&0.09&0.32&0.38&0.18\\
Answer Generation &0.27&0.20&0.27&0.19&0.20&0.24&0.20&0.12&0.19&0.10&0.29&0.38&0.24 \\
Answer Search &2.00&1.50&1.85&1.35&1.35&1.35&1.35&0.85&1.15&0.65&2.15&2.15& - \\
Translation & 1.05&-&-&-&-&-&-&-&0.55&-&-&-&-\\
\bottomrule
\end{tabular}
\caption{\label{tab:annotation_cost}
\small Annotation cost per unit for each task (in US dollars). The answer search task for English questions is annotated in-house so there is no external cost. The translation annotation is only conducted for German and Tamil.
}
\end{table*}

Helpful but not fully answering means the response contains helpful information, but is not enough to answer the question, or it can fully answers the question but the information is uncertain. “Helpful” means it provides useful information to help you know more about the question or narrow down the scope of the answer.

For example:
-question: Is it good for my 3-year-old kid?
-response: my 5-year-old son likes it.

It cannot fully tell whether a 3-year-old will like it, but knowing that a 5-year-old likes it is helpful information. It helps you narrow down the range of the answer — You know it is for kids but not adults, just not sure if it works exactly for 3-year-old.

“irrelevant” means the response provides zero useful information about the question, and is totally useless. Imagine you are a customer that raises this question, you should only select this option when you cannot find any useful information from the response.

\subsection{Answer Generation}
\label{app:ans_gen}
During the answer annotation, annotators are instructed to provide a natural, informative, and complete sentence to directly answer the user questions given the provided information in the response. The provided answer is required to be:
\begin{itemize}
    \item natural. It should be a fluent, natural-sounding sentence.
    \item informative. It should provide key information or explanations for users to better understand the question. It cannot be a single word like “Yes” or “No” without further content.
    \item complete. It should be a complete sentence that provides more context instead of a short span.
\end{itemize}
There is also a caveat to avoid copying the candidate exactly. Annotators should always extract useful information from it and show the reasoning step in the answer to make it a natural reply.
If the candidate is from a customer-provided content, they are further instructed to write from a third-party viewpoint. For user-provided contents, the answer will be in the form of ``A customer says he feels ...'' instead of ``I feel ...''. 

\subsection{Quality control and annotation cost}
\label{app:quality_cost}
Annotations are done through the centific platform~\footnote{\url{https://www.centific.com/}}. The whole annotation process is summarized in Figure~\ref{fig:anno_process}. From the Home of the webapp, we can see the status of the task (how many hits have been done and how many hits remain to be annotated). In the Quality Assessment mode, the assessor could search and select annotators and then check the completed hits at any time. When the assessor checks the hits, they can correct them directly and give feedback to the annotators, to improve annotation quality.
The annotation cost differs among languages and tasks. Table~\ref{tab:annotation_cost} provides a summary.

\begin{table*}
\centering
\small
\begin{tabular}{llll|cc|cccc}
\toprule
\multirow{2}{*}{\textbf{Language}} & \multirow{2}{*}{\textbf{Branch}} & \multirow{2}{*}{\textbf{Script}} & \multirow{2}{*}{\textbf{Market}} &\multicolumn{2}{c|}{\textbf{Train + Dev}}&\multicolumn{4}{c}{\textbf{Test}}
\\&&&&\textbf{\#Inst}&\textbf{\#Ans}&\textbf{\#Inst}&\textbf{\#Ans}&\textbf{\%Full}&\textbf{\%Rel}\\
\midrule
English (EN) & Germanic & Latin& US &131,520 &24,928 &20,142&4,392&84.1 &95.2\\
\midrule
German (DE) & Germanic & Latin& Germany & 5,110 & 806 & 10,201 & 1,504 &73.4 & 86.8 \\
Italian (IT) & Romance & Latin&Italy & 5,081 & 571 & 10,168 & 1,316 & 60.6 & 79.9\\
French (FR) & Romance & Latin&France & 5,047 & 838 & 10,135 & 1,684 & 71.1 & 96.7\\
Spanish (ES) & Romance & Latin&Spain & 5,055 & 1,003 & 10,112 & 1,961 & 78.5 & 91.5\\
Portuguese (PT) &  Romance & Latin&Brazil & 5,064 & 896 & 10,120 & 1,775 & 78.9 & 98.4\\
Polish (PL) & Balto-Slavic & Latin&Poland & 5,053 & 925 & 10,101 & 1,873 & 76.7 & 90.1\\
Arabic (AR) & Semitic & Arabic& SA & 5,097 & 752 & 10,178 & 1,544 & 71.3 & 84.6\\
Hindi (HI) & Indo-Aryan & Devanagari & India & 5,175 & 922 & 10,319 & 1,670 & 91.7 & 95.3\\
Tamil (TA) & Dravidian & Tamil & India & 5,076 & 892 & 10,166 & 1,584 & 73.4 & 81.7\\
Chinese (ZH) &  Sinitic & Chinese& China & 5,095 & 1,028 & 10,148 & 1,865 & 81.2 & 91.5\\
Japanese (JA) & Japonic & Kanji;Kana & Japan & 5,111 & 939 & 10,201 & 1,748 & 81.2 & 88.5\\
Korean (KO) &  Han &Hangul& US & 5,060 & 642 & 10,116 & 1,277 & 59.6 & 70.5\\
\bottomrule
\end{tabular}
\caption{\label{tab:data_stats}\small
Statistics of the ePQA and xPQA Datasets. \#Inst/\#Ans is the number of question-candidate pairs with relevance labels/manually written answers. \%Full/\%Rel is the percentage of questions that can be fully/partially answered.
}
\end{table*}

\subsection{Dataset Statistics}

To increase the number of negative samples, for every question we further randomly sample 5 candidates from the candidate set of corresponding products. These negative candidates, together with the annotated candidates, will form a closed-pool candidate set to evaluate the candidate ranker. Table~\ref{tab:data_stats} shows the statistics of the ePQA and xPQA datasets.

\section{Experiments}
\label{app:exp}
For the candidate ranking task, we initialize our model with Bert-base~\cite{devlin2019bert} in \textbf{translate-test} and mBert-base in the other two approaches. Following the common practice, we concatenate the question and candidate (split by the <SEP> token) and then feed it into the encoder. An MLP layer is added on top of the first <CLS> token to output three logits. These logits go through the softmax layer to represent the probability of three labels. At runtime, we use the probability of  ``fully answering'' as the score for each candidate.

For the answer generation task, we initialize our model with T5-base~\cite{raffel2020exploring} for the \textbf{translate-test} approach and mT5-base~\cite{xue2021mt5} for the other two approaches. The input is the question concatenated with the candidate and the output is the ground-truth answer. At runtime, we generate the output with beam search (beam size as 5). Both the ranker and generator are trained with standard cross entropy loss. 

We implement all models based on the Huggingface Transformers library~\footnote{\url{https://huggingface.co/}} with PyTorch~\footnote{\url{https://pytorch.org/}}. Models are optimized with the Adam optimizer~\cite{kingma2014adam}. We truncate the total input length to 128 subword tokens and select the learning rate from [5e-6, 1e-5, 3e-5, 5e-5, 1e-4]. The warm-up step is selected
from [5\%, 10\%, 20\%, 50\%] of the whole training
steps. For the ranker, we choose the
best configuration based on the accuracy of the validation set. For the generative model, we choose the
best configuration based on the perplexity of the
validation set. In the end, we set the learning rate
of the ranker as 3e-5 and that of the generator as
1e-5. The warm-up steps are set to 20\% for both. The batch size is set as 64. We evaluate the model performance every 1\% of the whole training step to select the best checkpoint. All
models are trained on one AWS P3.16 instance which includes 8 Nvidia V100 GPUs. The random seed is set as 42.
\end{document}